\documentclass[letterpaper, 10 pt, conference]{ieeeconf}  

\IEEEoverridecommandlockouts                              

\let\labelindent\relax
\usepackage[inline]{enumitem} 
\usepackage{graphicx}
\usepackage{subcaption}

\usepackage{epsfig} 
\usepackage{amsmath} 
\usepackage{amssymb}  

\usepackage{url}
\usepackage{color}

\usepackage{multirow}
\usepackage{siunitx}
\usepackage{makecell}

\makeatletter
\let\NAT@parse\undefined
\makeatother

\def\figref#1{Fig.~\ref{#1}}

\def\eqref#1{Eq.~(\ref{#1})}

\DeclareMathAlphabet\mathbfcal{OMS}{cmsy}{b}{n}
\usepackage[numbers]{natbib}
\newcolumntype{L}[1]{>{\raggedright\arraybackslash}p{#1}}
\newcolumntype{C}[1]{>{\centering\arraybackslash}p{#1}}
\newcolumntype{R}[1]{>{\raggedleft\arraybackslash}p{#1}}
\title{\LARGE \bf
Learning a Local Feature Descriptor for 3D LiDAR Scans
}

\author{Ayush Dewan \and Tim Caselitz \and Wolfram Burgard  
\thanks{All authors are with the Department of Computer Science at the University of Freiburg, Germany.}%
}

\begin{document}

\maketitle
\thispagestyle{empty}
\pagestyle{empty}

\begin{abstract}
Robust data association is necessary for virtually every SLAM system and finding
corresponding points is typically a preprocessing step for scan alignment
algorithms. Traditionally, handcrafted feature descriptors were used for these 
problems but recently learned descriptors have been shown to perform 
more robustly. In this work, we propose a local feature descriptor for 3D LiDAR 
scans. The descriptor is learned using a Convolutional Neural Network (CNN).
Our proposed architecture consists of a Siamese network for learning a feature descriptor and
a metric learning network for matching the descriptors. We also
present a method for estimating local surface patches and
obtaining ground-truth correspondences.  In extensive experiments,
we compare our learned feature descriptor with existing 3D local descriptors and report
highly competitive results for multiple experiments in terms of matching
accuracy and computation time.
\end{abstract}

\section{Introduction}

For many robotics tasks, it is required to
have robust data association in order to match same parts of the
scene under different conditions. Estimating data association is
always an important step in SLAM systems~\cite{serafin2016fast} and
different methods for lifelong visual localization~\cite{naseer14aaai}
rely on finding corresponding points between scenes captured in
different seasons. Furthermore, knowing corresponding points is also a
requirement for different scan alignment methods~\cite{dewan2016rigid}. 
In this work, we propose a local feature descriptor
for 3D LiDAR data. Our proposed Convolutional Neural Network (CNN)
architecture learns the feature descriptor and the metric for matching
the descriptor in a unified way. Additionally, we propose a method for 
generating local surface patches and discuss an approach for
obtaining ground-truth correspondences.


The majority of existing feature
descriptors for 3D data are handcrafted~\cite{alexandre20123d} and
rely either on quantifying surface normals or curvature around
keypoints. In contrast to these methods, we do not try to explicitly 
extract  geometric information but instead use raw scan data. In this work, we
focus on using LiDAR scans and learn descriptors for two channel surface
patches encoding local shape and surface reflectance values.
\begin{figure}[t]
  \centering
    \includegraphics[width=0.40\textwidth]{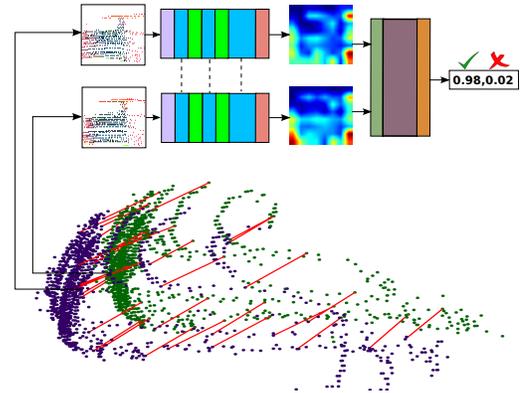}
    \caption{An illustration of keypoint matching using our learned
      feature descriptor. Surface patches around the keypoints are passed
      through the feature learning network to estimate feature descriptors.
      These descriptors are then passed through the metric learning network to estimate a
      matching score. Red lines show the correspondences between
      the keypoints in the two 3D LiDAR scans.
      Different colors in the architecture represent different
      layers, which are explained in later sections.
     }
    \label{fig:covergirl}
\end{figure}

Recently, various CNN-based methods have been proposed for learning feature descriptors for image
patches~\cite{han2015matchnet,zagoruyko2015learning,balntas2016pn} and
dense 3D surface patches~\cite{zeng20163dmatch}. All of these methods include a two or multi-stream Siamese network for learning a feature descriptor which is either discriminative in a predefined ~\cite{zagoruyko2015learning,balntas2016pn,zeng20163dmatch} or in a learned metric~\cite{han2015matchnet}.
Our proposed architecture consists of a two-stream Siamese network for learning a descriptor, 
followed by a network for learning a metric for matching the descriptor. Our feature learning network is based on the recent dense blocks architecture~\cite{huang2017densely}, and for metric learning, we use a stack of fully connected layers. Since the data used by existing learning based methods either consists of grayscale image patches or dense 3D surface patches we also generated our own training data. We extracted surface patches from LiDAR scans in the KITTI tracking
benchmark~\cite{Geiger2012CVPR} and obtained ground-truth correspondences by tracking keypoints using our previously proposed method~\cite{dewan2016rigid}. Fig.~\ref{fig:covergirl} demonstrates the matching of keypoints using our learned feature descriptor on sparse 3D LiDAR scans.

The foremost contribution of our work is a local feature descriptor
for  sparse 3D LiDAR data and a metric  for matching this feature descriptor.
We also target relevant problems
in the feature learning pipeline, i.e, extracting local surface patches and
obtaining the ground-truth correspondences. To validate the performance of our feature descriptor, 
we evaluate the matching accuracy and compare the performance with handcrafted feature descriptors and descriptors
learned with different architectures. We also present results for another experiment, where we 
align multiple objects based on feature correspondences. Furthermore, to highlight the difference between 
using a predefined and a learned metric, we present comparative results for our proposed feature descriptor
using respective cases. In addition, we also show that our descriptor can generalize to data from different type of LiDAR sensors.
We do this by repeating the alignment experiment but with data from a different sensor. We also present an ablation study to
provide insight on the role of each modality we use for learning the descriptor. The training and test data, learned models and a C++ API for using the feature descriptor with PCL is available here. \footnote{http://deep3d-descriptor.informatik.uni-freiburg.de}

\section{Related Work}

In this section we briefly discuss the handcrafted feature
descriptors that we use for comparison with our method and
an existing feature descriptor for sparse LiDAR scans.
We also discuss different CNN based descriptors proposed for grayscale and dense 3D surface patches.

Several handcrafted local feature descriptors for 3D pointclouds have
been proposed~\cite{alexandre20123d} and are currently part of the PCL
library~\cite{rusu20113d}. The first descriptor we compare with is
the Fast Point Feature Histogram (FPFH)~\cite{rusu2009fast}, which
requires normals as input and generalizes the mean
curvature around a point using a
histogram.~\citeauthor{tombari2010unique}~\cite{tombari2010unique}
proposed the Signatures of Histograms of Orientations (SHOT) descriptor. Their contribution is a method for
robustly estimating local reference frames and a descriptor that
quantifies local surface normal information. The third image
descriptor we compare with is 3D Shape Context
(3DSC)~\cite{frome2004recognizing}. Similar to other descriptors, it
also requires surface normals as input and uses a spherical grid around
the keypoint for counting the number of points in bins along azimuth,
elevation and radial coordinates.  We compare our results with these
descriptors and show the advantages of using a learned feature
descriptor over the handcrafted counterparts.

A feature descriptor designed for sparse
3D LiDAR scans was proposed by \citeauthor{serafin2016fast}~\cite{serafin2016fast}.
They quantify the vertical structure in the
scene with 3D lines and planes (circles) and show the efficacy of
their approach by integrating these descriptors in a SLAM system. Even
though their method improves the performance of the SLAM system, it
might perform sub-optimally for environments that lack these
vertical structures. In contrast to this, our feature descriptor is not quantifying any geometric structures in the environment and therefore can work in different environments. 

With the advent of CNNs, several methods have been proposed for
learning feature descriptors, but mainly for grayscale image patches. The
architectures discussed in these methods consist of a network for
learning the descriptor, followed by either a metric learning network for matching
the descriptors or a loss layer which minimizes the distance between
the descriptors using a predefined
metric. The architecture proposed in MatchNet~\cite{han2015matchnet}
consists of a Siamese network
for learning the descriptor and a metric learning
module. DeepCompare~\cite{zagoruyko2015learning} discusses several
different architectures: Siamese, pseudo-Siamese, two-channel and
central-surround two-stream networks.  The difference between Siamese
and pseudo-Siamese is that in the latter weights are shared only
for selected layers instead of every layer. In two-channel architectures, the image
patches are stacked as two channels instead of having a Siamese
network. Central-surround two-stream networks consist of four input
streams, two for complete image patches and two for the central crop
of the input patches. They use a loss layer minimizing Euclidean
distance instead of a metric learning module and show that the central-surround two-stream
architecture gives the best performance. MatchNet
and DeepCompare are outperformed by recently proposed PN-Net~\cite{balntas2016pn}. This
network architecture uses three input streams where two streams have
matching image patches and the third stream is a non-matching image
patch. Unlike MatcheNet, which uses softmax loss, they use
SoftPN loss, a modified version of hinge loss~\cite{zagoruyko2015learning}.

A learning-based approach for 3D data was recently proposed by
~\citeauthor{zeng20163dmatch}~\cite{zeng20163dmatch}. Their approach
called 3DMatch, targets learning descriptors for dense 3D surface patches
using a Siamese network trained with contrastive $l_2$ loss.
One of the main contributions is the proposed 3D patch representation, 
which uses a 3D voxel grid of Truncated Distance Function values 
for representing the 3D shape. The key difference between their work and
ours is that they target dense 3D surface patches extracted after
aligning multiple scans, whereas we focus on learning descriptors for
single sparse 3D LiDAR scan.

We compare our proposed feature descriptor with other feature descriptors known to work with sparse 3D LiDAR data. A comparison with other learned feature descriptors is not possible because these descriptors are learned for either grayscale 
image patches or dense 3D surface patches. Additionally, to justify using the dense blocks based architecture for learning, we present comparative results with descriptors learned using different types of CNN architectures~\cite{han2015matchnet,he2016deep}

%
%

\begin{figure}[t]
  \centering
  \includegraphics[width=0.35\textwidth]{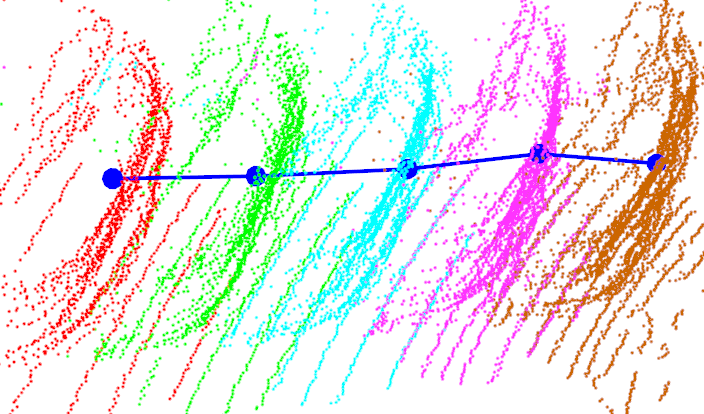}
  \includegraphics[width=0.35\textwidth]{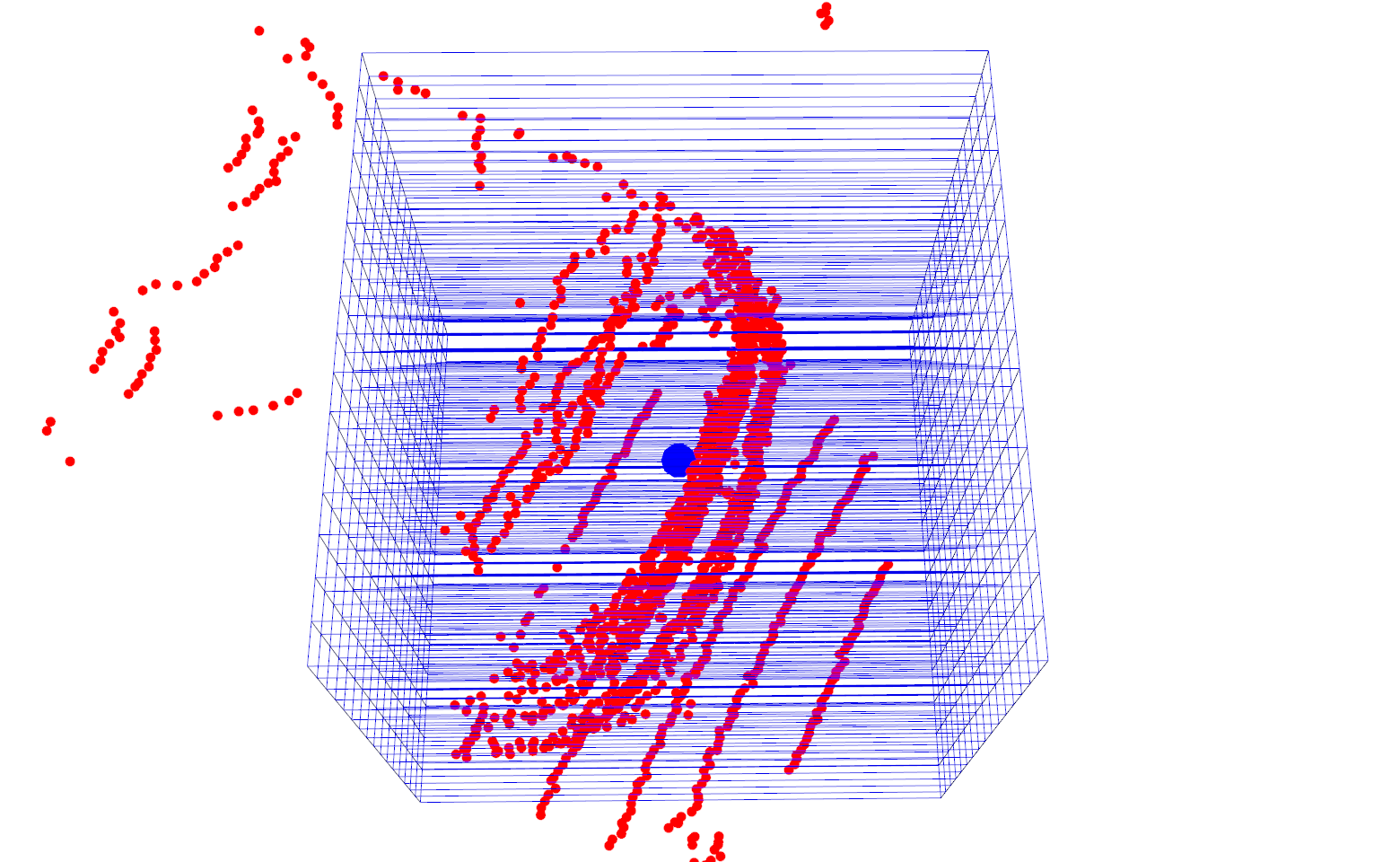}
  \includegraphics[width=0.23\textwidth]{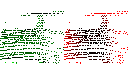}

    \caption{The top image shows the tracked keypoint for five frames. The 
image in the middle shows a voxelized cube around a keypoint.
The bottom image shows different channels of an extracted surface
patch. Surface reflectance intensity is showed in green and depth is shown in red.
}
    \label{fig:patch_generation}
\end{figure}

\section{Learning a Local Feature Descriptor}

Our feature learning pipeline has three main steps. First we
extract keypoints and track them to obtain ground-truth 
correspondences. Then we extract surface patches around
successfully  tracked keypoints  and in the last step we train the network for 
learning a feature descriptor as well as a metric for matching 
the descriptors in a unified manner.

\subsection{Generating Training Data}

A key requirement for supervised learning is labeled training
data. Since labeling many correspondences by hand is a strenuous task, existing methods~\cite{han2015matchnet,zagoruyko2015learning,balntas2016pn,zeng20163dmatch} 
use 3D scene reconstruction for associating pixels corresponding to same
3D point for obtaining ground-truth correspondences. Using datasets by these methods is not possible, since our objective is to learn a feature descriptor for sparse 3D LiDAR data and the datasets made available by these methods either consists of grayscale image patches or dense 3D surface patches.
\subsubsection{Ground-Truth Correspondences}
To obtain ground-truth correspondences, we first select the keypoints using uniform
sampling and then track those keypoints for the next five
frames. For tracking, we use our previously proposed method for estimating pointwise
motion~\cite{dewan2016rigid}. Associating keypoints over multiple frames instead of one
allows us to remove false correspondences. For keypoints that are
successfully tracked over multiple frames, we extract surface patches
for all five frames. The top image in
Fig.~\ref{fig:patch_generation} shows tracking of a keypoint for five frames.

We used the LiDAR scans from the KITTI tracking benchmark which consists of
20 sequences. We used the surface patches extracted from the first
10 sequences for training and the remaining 10 for testing.

\subsubsection{Training Patches}

Training a neural network for learning a feature descriptor requires local surface patches around keypoints. In case of 2D image data, generating patches is a straightforward task since the data is organized in a grid structure, but for unorganized sparse 3D pointclouds this task is non-trivial. In our approach, for a given keypoint, we
generate a cube with predefined length that is divided into $64\times64\times1$ voxels.  For every voxel we calculate the average distance w.r.t the keypoint and the  average surface reflectance intensity values for the points inside the voxel. and store the 3D voxel as a two channel image patch ($64\times64\times2$). The first modality (depth) aims at capturing the geometry and the second (intensity) captures surface reflectance properties. The local extent in which information has to be captured around a keypoint is defined by the length of the cube.
Fig.~\ref{fig:patch_generation} illustrates this process, where
the middle image shows the voxel structure around a keypoint and the
bottom image shows the modalities we use.
\begin{figure*}[t]

  \centering
	  \begin{subfigure}[t]{0.65\textwidth}
    
    \includegraphics[width=0.80\linewidth]{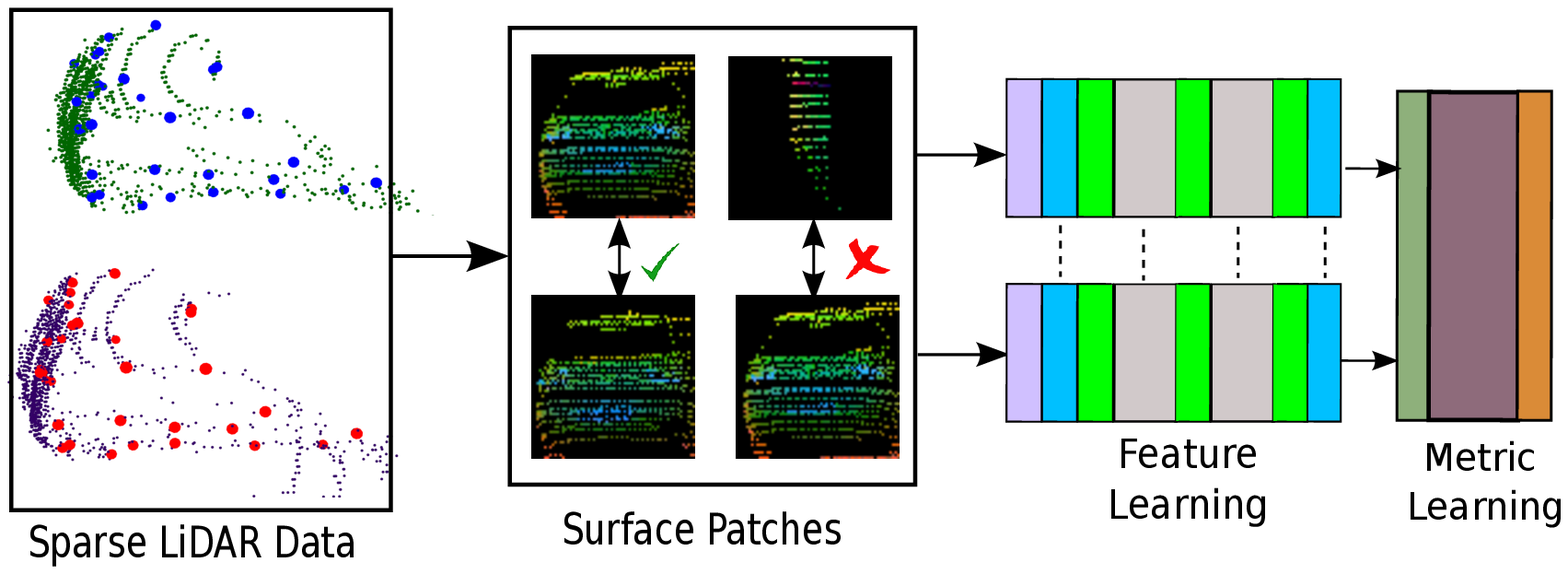}
    \caption{}
    \label{fig:summary}
  \end{subfigure}%
  \centering
  \begin{subfigure}[t]{0.35\textwidth}
    \centering
    \includegraphics[width=0.85\linewidth]{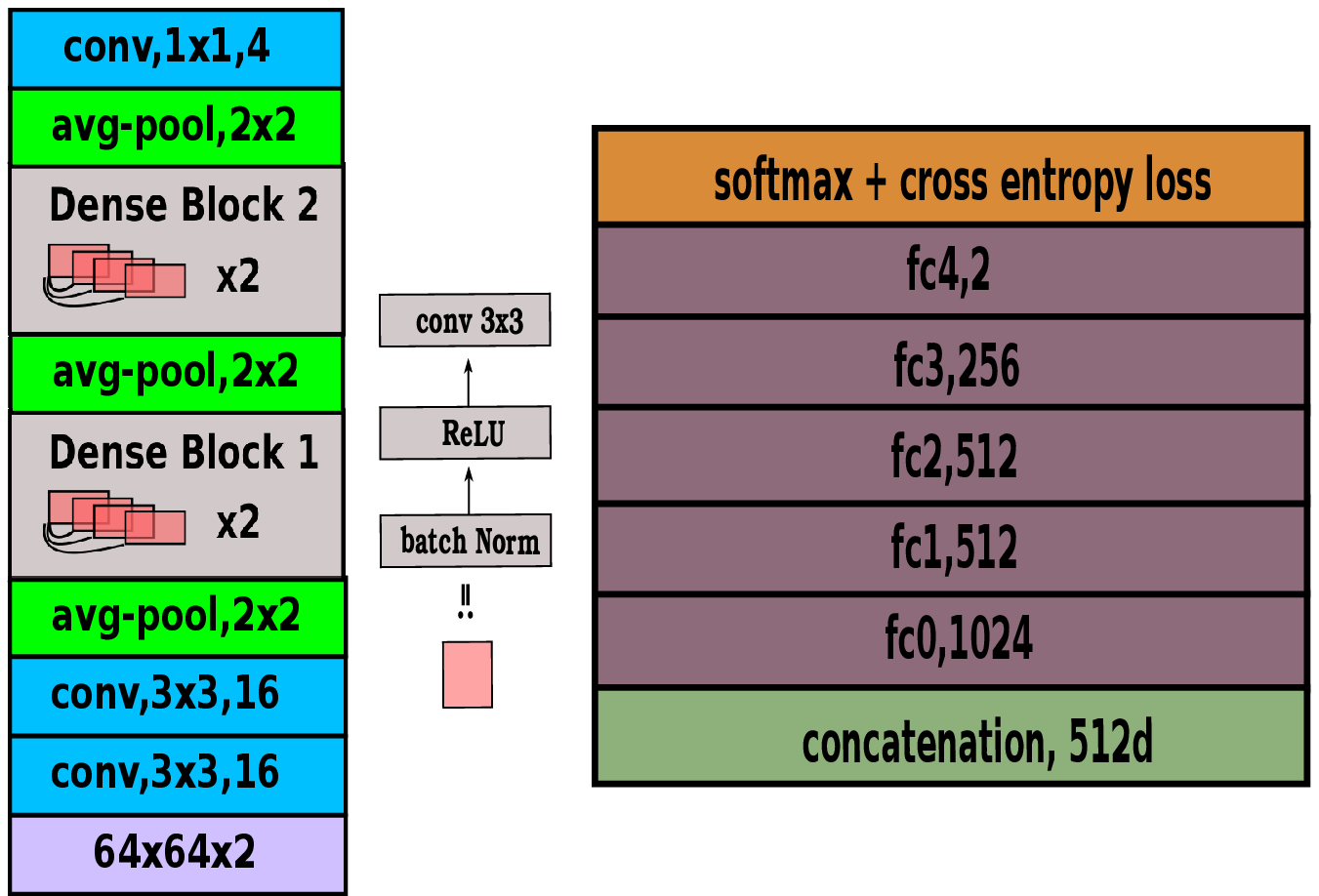}
    \caption{}
    \label{fig:architecture}
    
  \end{subfigure}%
    \caption{(a) An overview of our feature learning method. The local
      information around the keypoints from sparse LiDAR scans is
      converted into surface patches. Input to the feature learning network is a pair
      of matching and a pair of non-matching surface patches and input to the metric learning network is the learned feature
      descriptors. In (b) we show the architecture for each module. Feature learning module
      is a Siamese network, where filters in each layer share weights and the metric learning network contains a stack of
      fully connected layers.}
    
\end{figure*}
\subsection{Network Architecture}\label{sec:architecture}

Fig.~\ref{fig:summary} shows the network architecture that we use in
our approach. The architecture consists of a two-stream Siamese
network for learning the features, followed by a metric learning network.
Each stream of the Siamese network consists of two convolution layers followed
by two dense blocks and a \textit{bottleneck} layer. Dense blocks based networks~\cite{huang2017densely} have been recently shown to improve the state-of-the-art for different tasks. According to the authors~\cite{huang2017densely}, the improvement in performance is mainly attributed to the  better exploitation of feature re-use capability which allows efficient flow of information resulting in better feature representation. 

Each dense block consists of two layers, where each layer is a composite function consisting of batch-normalization, Rectified Linear Unit (ReLU), and a convolution operation. The \textit{bottleneck} layer is a convolution layer and output of it is the learned feature descriptor.

In the metric learning module, we have 5 fully connected layers
(\textit{fc}) where every \textit{fc} layer, except the last one
(\textit{fc4}), is followed by a ReLU. The input to the metric learning
module is the concatenation of the output of each stream of the
Siamese network. The number of feature maps for
every \textit{fc} layer is shown in Fig.~\ref{fig:architecture}.

Our unified feature and metric learning problem can be seen as a
binary classification problem, where a pair of input patches have to
be classified as \textit{matching} or \textit{non-matching}. Our
training set is $\mathbfcal{T} = \{(\mathit{X}^1_{n},\mathit{X}^2_{n}, 
\mathit{Y}_{n}),\;n=1,\dots, N\}$,
where $\mathit{X}^1_{n}$ and $\mathit{X}^2_{n}$ are two sets of surface patches
and $\mathit{Y}_{n} = \lbrace y_k\in\lbrace1,0\rbrace,\;k=1,\dots,
N\rbrace$ is the corresponding ground truth labels. The activation function for our learning
model is defined as $f(x^1_k,x^2_k,\theta)$, where $\theta$ are the
parameters of our model and $x^1_k\in\mathit{X}^1_n$ and
$x^2_k\in\mathit{X}^2_n$ is a pair of surface patches. The network
learns the weights $\theta$ by minimizing the cross-entropy
(softmax) loss in Eq.~\ref{eq:entopyloss}, over all patch
pairs as shown in Eq.~\ref{eq:deeploss}. To avoid overfitting we add
a term for $l_2$ regularization to the loss function in
Eq.~\ref{eq:deeploss}.
\begin{align} 
\mathbfcal{L}(p,q) &= -\sum\limits_{c\in\{0,1\}} p_{c} \log{q_{c}} \label{eq:entopyloss} \\
\theta^{*} = \underset{\theta}{\operatorname{argmin}}\frac{1}{N}
\sum_{k=1}^{N} &\mathbfcal{L}\left (y_k , f\left (x^1_k,x^2_k,\theta\right ) \right ) +
\frac{\eta}{2}\Vert\theta\Vert^2_2\label{eq:deeploss}
\end{align}

An alternative to metric learning is to use a loss layer directly
after the feature learning. Most
commonly used loss layers are contrastive loss~\cite{zeng20163dmatch} and
hinge embedding loss~\cite{simo2015discriminative}. Both of these layers try to minimize
the Euclidean distance between matching descriptor pairs and simultaneously increase the \textit{margin} between non-matching pairs. In section~\ref{sec:results} we present a comparison between our proposed network for unified learning and our modified feature learning network trained with hinge embedding loss. In the modified version we replace the last convolution layer (Fig.~\ref{fig:architecture}) with a set of fully connected layers, where the output of the last layer is the feature descriptor.

\subsection{Training}

Our complete network architecture is implemented in
TensorFlow~\cite{abadi2016tensorflow}. Using our patch generation
method, we generated $58,\!710$ surface patches for training. The input
to the network is a batch of surface patch pairs. Each batch consists
of an equal number of matching and non-matching surface patches as shown
in Fig.~\ref{fig:summary}. With these surface patches, we estimated
$117,\!400$ positive and $704,\!400$ negative pairs. Since the input to the
network always consists of a negative
and positive combination, our effective training set consists of
$704,\!400$ samples.

We train our network with a batch size of $32$ and use the Adam
optimizer~\cite{kingma2014adam} with a learning rate of $1e^{-4}$. The
parameter $\eta$ for $l_2$-norm regularization was fixed to
$5e^{-4}$. The growth rate for dense blocks is 4. 
The network was trained for 5 epochs and the complete
training process required around 2 hours on an NVIDIA GeForce GTX 980
graphics card.

\section{Results}\label{sec:results}

\begin{figure}[t]
  \centering
  \includegraphics[width=0.30\textwidth]{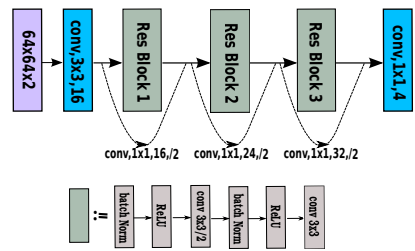}
    \caption{Architecture for ResNet-8}
    \label{fig:resnet}
\end{figure}

To evaluate our method, we perform multiple experiments. We first evaluate the matching accuracy for various descriptors. We then report the average alignment errors for objects scanned using Velodyne HDL-64E and HDL-32E LiDAR scanners. We also report computation times for calculating and matching the different descriptors. For all experiments we compare the proposed feature descriptors with other feature descriptors. Among handcrafted descriptors, we compare with SHOT~\cite{tombari2010unique}, 
FPFH~\cite{rusu2009fast}, and 3DSC~\cite{frome2004recognizing}. To justify the usage of dense blocks for our task, we present results for feature descriptors learned with the following two different network architectures~\cite{han2015matchnet,he2016deep}. 
\begin{enumerate}
\item The first architecture~\cite{han2015matchnet} was proposed to learn a feature descriptor and a metric for grayscale image patches. Their feature learning architecture consists of blocks of convolution layers and ReLUs separated by max-pooling layers and they use a stack of fully connected layers for metric learning. This feature learning architecture is similar to the initially proposed CNN architectures for instance VGG~\cite{Simonyan14c}. Other methods~\cite{zeng20163dmatch,zagoruyko2015learning} proposed for learning feature descriptors have also used similar architectures. The reason we chose the architecture from MatchNet is because they have similar input patch size as ours $(64\times 64)$ and also use metric learning. 
\item The second architecture we compare with consists of \textit{residual} blocks~\cite{he2016deep}. Lately, residual blocks based architectures have also shown to perform well for a variety of tasks~\cite{he2016deep,Loquercio_2018}. We use a ResNet-8~\cite{Loquercio_2018} for feature learning and keep the metric learning architecture unchanged.~\figref{fig:resnet} shows the ResNet-8 architecture we used. 
\end{enumerate}
Furthermore, we also compare with our feature learning network
trained with the hinge embedding (H.E.) loss. The first  three comparisons 
present the advantages of learned feature descriptors 
over the handcrafted descriptors. The next two comparisons highlight the 
benefits of using dense blocks while the last one compares the performance 
of a learned metric with a predefined metric. 

\subsection{Matching Accuracy}
\begin{figure}[t]
  \centering
  \includegraphics[width=0.42\textwidth]{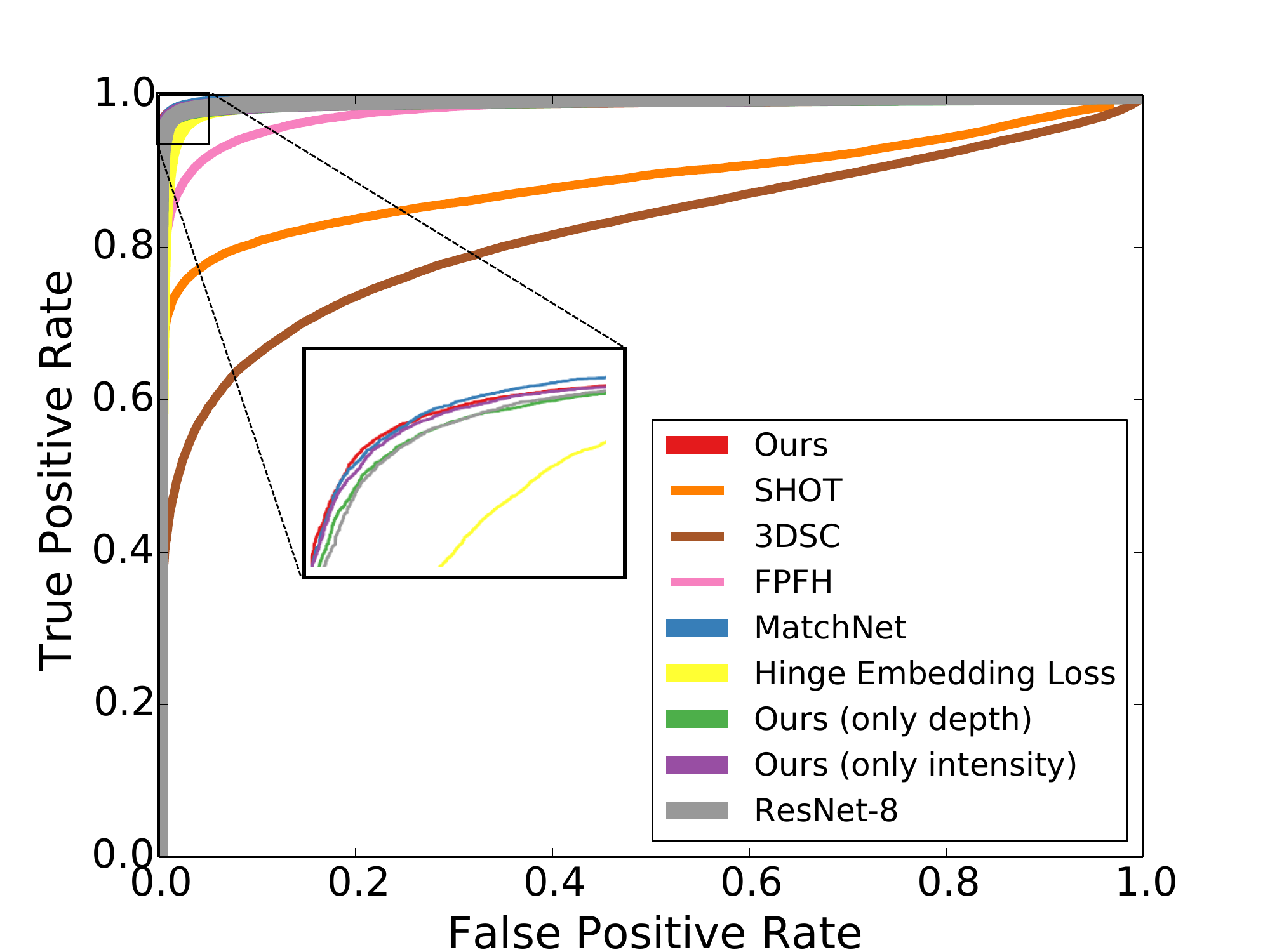}
    \caption{ROC curves for different feature descriptors. Performance
      of our proposed feature descriptor is comparable to
      MatchNet. Using H.E. loss instead of metric learning leads to
      a decrease in performance, whereas all handcrafted descriptors
      underperform in comparison to the learned descriptors.}
    \label{fig:roc_curve}
\end{figure}

The goal of this experiment is to test the performance of different descriptors on the pairs of surface patches from our testing set. We use $100,\!000$ samples, half of them are matching and the other half are non-matching. For every case we plot a receiver operating characteristic (ROC) curve and report false-positive rate at 95\% recall (FPR95). The ROC curves are shown in Fig.~\ref{fig:roc_curve} and the FPR95 is reported in Table~\ref{tab:FPR}. The curve is plotted for various matching thresholds, which in case of metric learning is the softmax score. For the handcrafted descriptors  and the descriptor learned using H.E. loss, the Euclidean distance between descriptors is used as matching threshold.

The error for our feature descriptor is the lowest and it outperforms the handcrafted descriptors
by a significant margin. The next best performing feature descriptors is learned using the
architecture of MatchNet, followed by the ResNet-8 architecture. These results justify learning a feature descriptor using our proposed architecture. 

The error increases when our feature learning network is
trained with H.E. loss (yellow curve in
Fig.~\ref{fig:roc_curve}), demonstrating the importance of
metric learning. The increase in performance due to metric learning
comes at the cost of an increase in matching time due to
computationally expensive forward pass through
the metric learning network. We discuss the matching time for different descriptors in next section.
 \begin{figure*}[t]
  \centering
  \includegraphics[width=0.40\textwidth]{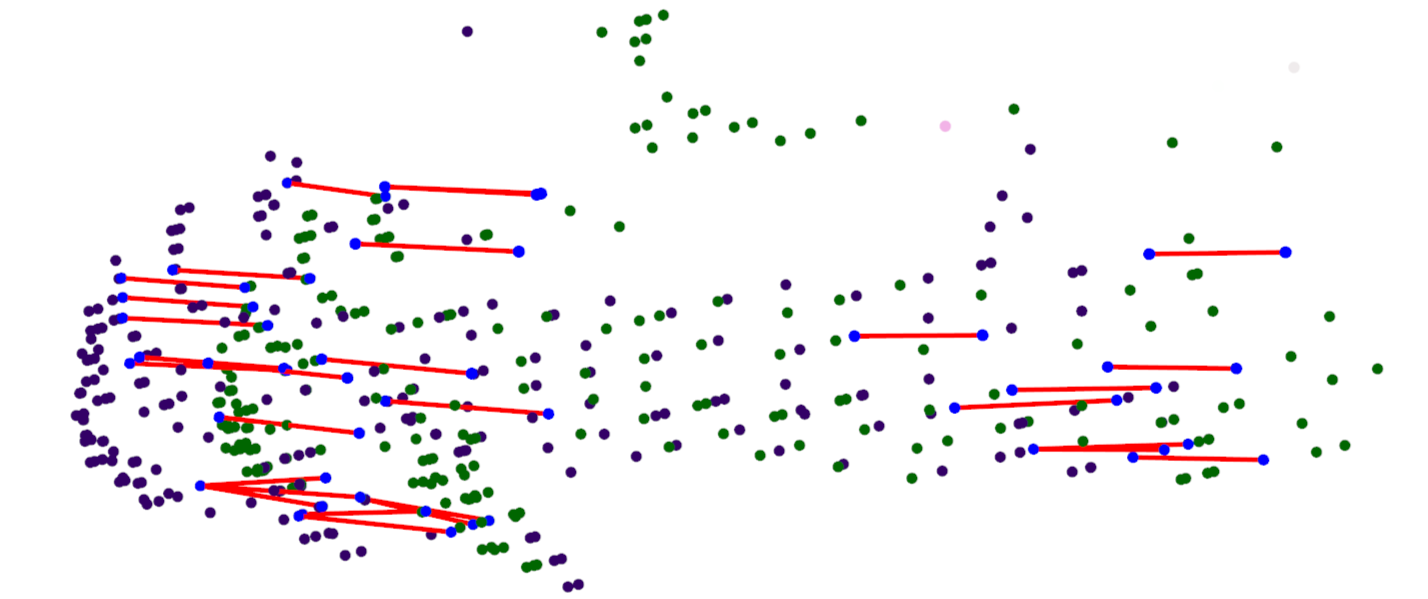}
  \includegraphics[width=0.40\textwidth]{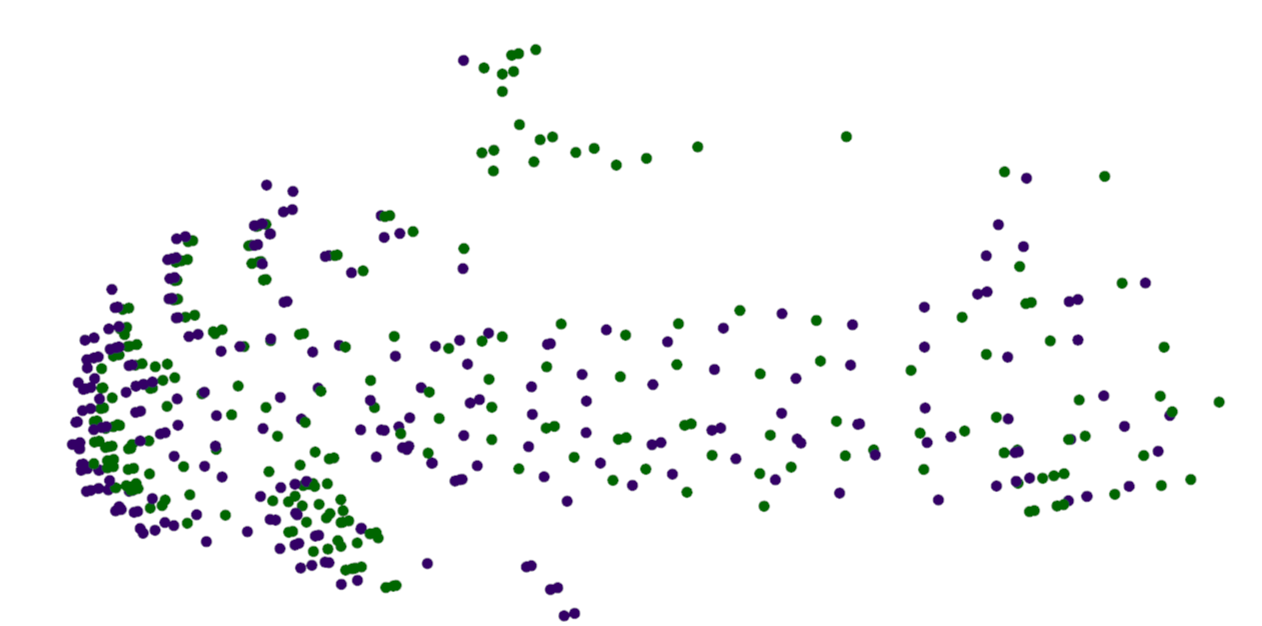}

    \caption{An illustration of the alignment experiment. The image on
      the left shows the sparse misaligned objects and the
      correspondences estimated by matching our feature
      descriptor. The image on the right shows the aligned pointclouds.}
    \label{fig:lidar-alignment}
\end{figure*}
\begin{table}[h!]
 \centering
 \caption{FPR95 Error}
 \begin{tabular}{ |c|C{1.8cm}|C{1.8cm}|}
 \hline
 Method&Feature Size& Error(\%)\\\hline
 SHOT~\cite{tombari2010unique}&\hspace{0.15cm}352&82.56\\
 FPFH~\cite{rusu2009fast}&\hspace{0.3cm}33&10.26\\
 3DSC~\cite{frome2004recognizing}&1980&89.16\\
 MatchNet~\cite{han2015matchnet}&4096&\hspace{0.15cm}0.45\\
 ResNet-8&\hspace{0.15cm}256&\hspace{0.15cm}0.60\\
 Ours (H.E. Loss)&\hspace{0.15cm}256&\hspace{0.15cm}1.94\\
 Ours (depth + intensity) &\hspace{0.15cm}256&\hspace{0.15cm}\textbf{0.42}\\
 Ours (only depth)&\hspace{0.15cm}256&\hspace{0.15cm}0.46\\
 Ours (only intensity)&\hspace{0.15cm}256&\hspace{0.15cm}0.53\\
 \hline
 \end{tabular}
  \label{tab:FPR}
\end{table}
\begin{table}[h!]
 \centering
 \caption{Alignment error of our method for the individual objects}
 \begin{tabular}{ |c|C{1.4cm}|C{1.4cm}|C{1.4cm}|C{1.4cm}|}
 \hline

 Object ID&Points&
  Keypoints&$t_e(\si{\meter})$&$r_e(\si{\radian})$\\\hline
  \multicolumn{5}{|c|}{Scans from Velodyne HDL-64E}\\\hline
  0&1369&483&0.28&0.04\\
  1&\hspace{0.15cm}493&285&0.54&0.04\\
  2&\hspace{0.15cm}787&332&0.47&0.04\\
  3&\hspace{0.15cm}250&186&0.52&0.02\\
  4&1320&383&1.55&0.24\\
  5&\hspace{0.15cm}970&394&1.06&0.13\\
  6&\hspace{0.15cm}228&129&1.48&0.10\\
  7&\hspace{0.15cm}199&154&1.10&0.06\\
  8&\hspace{0.15cm}580&395&0.73&0.02\\
  9&\hspace{0.15cm}564&427&0.08&0.02\\
  10&\hspace{0.15cm}316&233&1.22&0.04\\
  11&1517&908&0.23&0.01\\\hline
  \multicolumn{5}{|c|}{Scans from Velodyne HDL-32E}\\\hline
  0&\hspace{0.15cm}313&\hspace{0.15cm}230&0.88&0.07\\
  1&3271&1099&0.46&0.13\\
  2&3741&1239&0.49&0.13\\
  3&\hspace{0.15cm}319&\hspace{0.15cm}239&0.22&0.14\\  
 \hline
 \end{tabular}
  \label{tab:alignment_experiment_details}
\end{table}
\begin{table}[h!]
 \centering
 \caption{Average alignment errors for HDL-64E scans}
 \begin{tabular}{|C{1.47cm}|C{1.2cm}|C{1.5cm}|C{1.25cm}|C{1.25cm}|C{1.0cm}|}
 \hline
 \multirow{2}{*}{Method}&\multicolumn{2}{ c |}{raw}
 &\multicolumn{2}{ c |}{RANSAC}&\multirow{2}{*}{t(s)}\\\cline{2-5}
 &$t_e(\si{\meter})$&$r_e(\si{\radian})$&$t_e(\si{\meter})$&$r_e(\si{\radian})$&\\
 \hline
 
  SHOT~\cite{tombari2010unique}&1.38$\pm$1.06&0.13$\pm$0.11&0.76$\pm$0.56&0.10$\pm$0.07&\hspace{0.15cm}0.05\\
  FPFH~\cite{rusu2009fast}&3.37$\pm$5.40&0.33$\pm$0.54&1.30$\pm$0.89&0.14$\pm$0.11&\hspace{0.3cm}\textbf{0.004}\\
  3DSC~\cite{frome2004recognizing}&2.75$\pm$2.58&0.34$\pm$0.38&0.81$\pm$0.37&0.08$\pm$0.04&\hspace{0.15cm}0.57\\
  MatchNet~\cite{han2015matchnet}&0.88$\pm$0.47 &0.08$\pm$0.06
  &0.93$\pm$0.77&0.07$\pm$0.05&14.78\\
  ResNet-8&0.95$\pm$0.48 &0.077$\pm$0.067&0.83$\pm$0.71&0.08$\pm$0.05&\hspace{0.15cm}1.03\\
  H.E. loss&\textbf{0.76}$\pm$\textbf{0.34}&0.076$\pm$0.062&\textbf{0.52}$\pm$\textbf{0.27}&\textbf{0.05}$\pm$\textbf{0.01}&
 \hspace{0.15cm}0.18\\ 
  Ours&0.77$\pm$0.47 &\textbf{0.071}$\pm$\textbf{0.061}
  &0.57$\pm$0.28&0.06$\pm$0.02&\hspace{0.15cm}0.92\\
  \hline
 \end{tabular}
  \label{tab:alignment}
\end{table}
\begin{table}[h!]
 \centering
 \caption{Average alignment errors for HDL-32E scans}
 \begin{tabular}{|c|C{1.4cm}|C{1.4cm}|C{1.4cm}|C{1.4cm}|}
 \hline
 \multirow{2}{*}{Method} &\multicolumn{2}{ c |}{raw}
 &\multicolumn{2}{ c |}{RANSAC}\\\cline{2-5}
 &$t_e(\si{\meter})$&$r_e(\si{\radian})$&$t_e(\si{\meter})$&$r_e(\si{\radian})$\\
 \hline
 
  SHOT~\cite{tombari2010unique}&0.57$\pm$0.29&0.11$\pm$0.04&0.46$\pm$0.54&0.07$\pm$0.06\\
  FPFH~\cite{rusu2009fast}&1.32$\pm$1.08&0.22$\pm$0.10&2.07$\pm$2.20&0.14$\pm$0.25\\
  3DSC~\cite{frome2004recognizing}&2.48$\pm$3.46&0.28$\pm$0.27&0.73$\pm$0.77&0.06$\pm$0.03\\
  MatchNet~\cite{han2015matchnet}&0.97$\pm$1.01 &0.13$\pm$0.08
  &0.88$\pm$1.24&0.14$\pm$0.11\\
  ResNet-8&0.81$\pm$0.33 &0.14$\pm$0.06&0.60$\pm$0.56&0.07$\pm$0.04\\
  H.E. loss&0.65$\pm$0.14 &\textbf{0.07}$\pm$\textbf{0.06}&0.58$\pm$0.55&0.11$\pm$0.04\\ 
  Ours&\textbf{0.51}$\pm$\textbf{0.23} &0.12$\pm$0.02
  &\textbf{0.41}$\pm$\textbf{0.39}&\textbf{0.05}$\pm$\textbf{0.05}\\
  \hline
 \end{tabular}
  \label{tab:alignment_32}
\end{table}

\subsection{Alignment}

Many methods for surface or scan registration require coarse initial alignment, especially when data is collected
at a low rate (typically 10Hz for LiDAR) and the assumption that nearest neighbor
points are corresponding does not hold. In this experiment we align multiple objects scanned using two different LiDAR scanners.
In Table~\ref{tab:alignment_experiment_details}, we report the number of objects used in the experiment, number of points belonging to each object, number of keypoints and the translational and rotational alignment error of our method.
\subsubsection{LiDAR Scans from Velodyne HDL-64E}\label{sec:alignment_64}
In this experiment we align 12 static objects extracted from consecutive LiDAR scans from the KITTI tracking benchmark. Since all objects are static, the ground-truth motion is the inverse sensor motion, which is provided by the benchmark. In Fig.~\ref{fig:lidar-alignment} we show example results, where the image on the left shows the misaligned objects and the corresponding points, whereas the right image illustrates the aligned pointclouds. As before, we use uniform sampling for selecting keypoints. In Table~\ref{tab:alignment}, we report the average translational error $t_e$ and rotational error $r_e$ for the motion estimated using the raw point correspondences and correspondences filtered with RANSAC. The motion estimated using raw correspondences reflects more clearly on the feature matching accuracy in comparison to filtered correspondences. The alignment error reported for our method using raw correspondences is average of the errors reported in Table~\ref{tab:alignment_experiment_details}. While our feature descriptors with learned metric and hinge embedding loss perform similar, they outperform the other descriptors by significant margin. 

The rightmost column in Table~\ref{tab:alignment} reports the feature matching time. The FPFH descriptor takes minimum time for matching but has the largest alignment error. All the handcrafted descriptors can be matched quickly using KD-trees in comparison to feature descriptors learned using a metric and therefore they have the lowest matching time. In this experiment our feature descriptor learned using Euclidean distance performs most favorably considering both alignment error and matching time. Among the feature descriptors learned using the metric, the time for MatchNet is the largest. This increase in time is mainly attributed to the large descriptor size (4096 vs.\ 256) which results in twice as much as parameters in the metric learning network in comparison to other learned descriptors. The matching time of ResNet-8 is similar to ours, since the architecture for metric learning and the feature descriptor size is the same in both cases. 
 
\subsubsection{LiDAR Scans from Velodyne HDL-32E}
We repeat the alignment experiment like above but with the data collected from a 32 beam LiDAR 
scanner. Like before we align static objects and use the sensor pose from the SLAM solution~\cite{kummerle2015autonomous} as 
ground-truth motion. The purpose of this experiment is to show that our feature descriptor is not overfitting to data collected from a single sensor but can generalize to data collected from different sensors. Even though both sensors provide the same modalities, the data from Velodyne HDL-32E is often sparser in comparison to data from Velodyne HDL-64E and also the has different measurement noise. Table~\ref{tab:alignment_32} shows the alignment error for different cases and our feature descriptor learned with metric outperforms the other descriptors. This experiment demonstrates that our proposed feature descriptor is capable of generalizing on data from different sensors.
 
 \begin{table}[h!]
 \centering
 \caption{Per feature computation time (in ms) for various neighborhood radii.}
\begin{tabular}{|c|C{1.2cm}|C{0.95cm}|C{0.95cm}|C{0.95cm}|C{1.1cm}|C{1.25cm}|}
\hline
\multirow{2}{*}{Method} & \multirow{2}{*}{Processor}&\multicolumn{5}{ c |}{Neighborhood radius (m)}\\\cline{3-7}
 & & 0.4 & 0.8 & 1.6&3.2&6.4 \\
 \hline
 SHOT~\cite{tombari2010unique}&MC&0.092&\textbf{0.096}&\textbf{0.119}&\hspace{0.15cm}0.284&\hspace{0.30cm}0.787\\
 FPFH~\cite{rusu2009fast}&MC&\textbf{0.090}&0.282&0.961&3.01&10.23\\
 3DSC~\cite{frome2004recognizing}&C&0.148&0.440&\hspace{-0.15cm}3.21&\hspace{-0.15cm}31.19&\hspace{-0.15cm}344.10\\
 MatchNet~\cite{han2015matchnet}&\makecell{MC+\\G}&\hspace{0.15cm}\makecell{0.036+\\\hspace{-0.15cm}0.139}&
 \hspace{0.15cm}\makecell{0.037+\\\hspace{-0.15cm}0.139}&\hspace{0.15cm}\makecell{0.047+\\\hspace{-0.15cm}0.139}
 &\hspace{0.30cm}\makecell{0.075+\\\hspace{-0.15cm}0.139}&\hspace{0.45cm}\makecell{0.219+\\\hspace{-0.15cm}0.139}\\
 ResNet-8&\makecell{MC+\\G}&\hspace{0.15cm}\makecell{0.036+\\\hspace{-0.15cm}0.151}&
 \hspace{0.15cm}\makecell{0.037+\\\hspace{-0.15cm}0.151}&\hspace{0.15cm}\makecell{0.047+\\\hspace{-0.15cm}0.151}
 &\hspace{0.30cm}\makecell{0.075+\\\hspace{-0.15cm}0.151}&\hspace{0.45cm}\makecell{0.219+\\\hspace{-0.15cm}0.151}\\
 H.E.
 loss&\makecell{MC+\\G}&\hspace{0.15cm}\makecell{0.036+\\\hspace{-0.15cm}0.132}&
 \hspace{0.15cm}\makecell{0.037+\\\hspace{-0.15cm}0.132}&\hspace{0.15cm}\makecell{0.047+\\\hspace{-0.15cm}0.132}
 &\hspace{0.30cm}\makecell{\textbf{0.075+}\\\hspace{-0.15cm}\textbf{0.132}}&
 \hspace{0.45cm}\makecell{\textbf{0.219+}\\\hspace{-0.15cm}\textbf{0.132}}\\ 
 Ours&\makecell{MC+\\G}&\hspace{0.15cm}\makecell{0.036+\\\hspace{-0.15cm}0.133}
 &\hspace{0.15cm}\makecell{0.037+\\\hspace{-0.15cm}0.133}&\hspace{0.15cm}\makecell{0.047+\\\hspace{-0.15cm}0.133}
 &\hspace{0.30cm}\makecell{0.075+\\\hspace{-0.15cm}0.133}&\hspace{0.45cm}\makecell{0.219+\\\hspace{-0.15cm}0.133}\\
 
 \hline
\end{tabular}
  \label{tab:neighbour_table}
\end{table}
\begin{table}[h!]
 \centering
 \caption{Computation time (in seconds) for various sampling radii.}
\begin{tabular}{|C{1.5cm}|C{0.98cm}|C{0.98cm}|C{0.98cm}|C{0.98cm}|C{0.9cm}|C{0.9cm}|C{0.9cm}|}
\hline
\multirow{2}{*}{Method}&\multicolumn{7}{ c |}{Sampling radius (m)}\\\cline{2-8}
 &3.2 & 1.6 & 0.8&0.4&0.2&0.1 &0.05 \\\hline
 SHOT~\cite{tombari2010unique}&\hspace{0.15cm}0.97&\hspace{0.15cm}0.97&\hspace{0.15cm}1.36
 &\hspace{0.15cm}2.64&6.45&12.47&24.83\\
 FPFH~\cite{rusu2009fast}&27.97&28.30&27.86&31.87&40.74&55.79&85.13\\
 3DSC~\cite{frome2004recognizing}&\hspace{0.15cm}8.97&35.07&99.49&355.58&1111.65&3073.30&6179.10\\
 MatchNet~\cite{han2015matchnet}&\hspace{0.30cm}\makecell{0.03+\\\hspace{-0.15cm}0.22}
 &\hspace{0.30cm}\makecell{0.11+\\\hspace{-0.15cm}0.33}
 &\hspace{0.30cm}\makecell{\textbf{0.28+}\\\hspace{-0.15cm}\textbf{0.67}}
 &\hspace{0.30cm}\makecell{0.81+\\\hspace{-0.15cm}1.68}
 &\makecell{3.72+\\3.29}
 &\makecell{4.40+\\6.25}
 &\makecell{9.17+\\8.26}\\
 ResNet-8&\hspace{0.30cm}\makecell{\textbf{0.03}+\\\hspace{-0.15cm}\textbf{0.20}}
 &\hspace{0.30cm}\makecell{\textbf{0.11+}\\\hspace{-0.15cm}\textbf{0.16}}
 &\hspace{0.30cm}\makecell{0.28+\\\hspace{-0.15cm}0.70} &\hspace{0.30cm}\makecell{0.81+\\\hspace{-0.15cm}1.68}
 &\makecell{3.72+\\3.61}
 &\makecell{4.40+\\6.75}
 &\makecell{9.17+\\9.84}\\ 
 H.E. loss&\hspace{0.30cm}\makecell{0.03+\\\hspace{-0.15cm}0.33}&\hspace{0.30cm}\makecell{0.11+\\\hspace{-0.15cm}0.42}
 &\hspace{0.30cm}\makecell{0.28+\\\hspace{-0.15cm}0.70}   
 &\hspace{0.30cm}\makecell{\textbf{0.81+}\\\hspace{-0.15cm}\textbf{1.47}}
 &\makecell{\textbf{3.72+}\\\textbf{3.01}}
 &\makecell{\textbf{4.40+}\\\textbf{5.43}}
 &\makecell{\textbf{9.17+}\\\textbf{7.80}}\\
 Ours&\hspace{0.30cm}\makecell{0.03+\\\hspace{-0.15cm}0.34}&\hspace{0.30cm}\makecell{0.11+\\\hspace{-0.15cm}0.43}
 &\hspace{0.30cm}\makecell{0.28+\\\hspace{-0.15cm}0.71}   
 &\hspace{0.30cm}\makecell{0.81+\\\hspace{-0.15cm}1.49}
 &\makecell{3.72+\\3.05}
 &\makecell{4.40+\\5.52}
 &\makecell{9.17+\\7.90}\\
 \hline
 
\end{tabular}

  \label{tab:sampling_table}
\end{table}

\subsection{Computation Time}
In Table~\ref{tab:neighbour_table} and~\ref{tab:sampling_table}, we report the computation time for estimating feature descriptors for different neighborhood radii and sampling radii. In the first case, we estimate descriptors for the same number of points (sampling radius of $0.4\si{\meter}$) but for different neighborhood radii. This evaluation highlights per feature calculation time, which only depends on the input neighborhood radius. In the second case, we use a fixed neighborhood radius of $3.2\si{\meter}$ while varying the sampling radius. This evaluation focuses on changes in computation time with the increase in number of keypoints. For each case we report the processor details, i.e whether it is a multi-core CPU (MC), single core CPU (C) or GPU (G). For learning methods, we separately report the time required for estimating the surface patches and the time required for estimating the feature descriptors. 

For small neighborhood radii, handcrafted descriptors have low computation time because different per-point operation like estimating normals and estimating descriptor among other different operations, are performed efficiently for smaller radii (fewer points). While increasing these radii (more points), these operations are not as efficient as before even with usage of KD-Trees. In case of the learned feature descriptors, the change in neighborhood radii only affects the patch computation time and not the feature estimation time. For patch computation the voxelization of the neighborhood is independent of the number of points in the neighborhood. The only operation dependent on number of points is calculating the average in depth and intensity values. Among the learned descriptors, our feature descriptor requires least computation time in comparison to other architectures because our feature learning network has the least number of parameters as well.

With the decrease in sampling radius, the number of keypoints increases. The reported time is the combined computation time for all the keypoints. In this case, the better performance of learned feature descriptors is mainly attributed to the proper utilization of parallel processing capabilities of GPUs in comparison to the multi-core implementation of handcrafted descriptors on CPUs. In comparison to other learned descriptors, the performance of our feature learning network scales better with the increase in number of keypoints. Since our network is smaller than others, it allows us to process larger batches of data in parallel.

\subsection{Ablation Study}

To better understand the contribution of each of the modality, we trained two separate networks with single channel 
input (depth and intensity). Table~\ref{tab:FPR} shows the FPR95 error for both cases. The performance of the feature descriptor learned using
depth is better than the feature descriptor learned using surface intensity values. Since the surface reflectance values often depend on the angle at which laser beam hits the surface, they are not as stable as the depth values, especially in the case when the sensor is moving. The error for both of these cases is higher than the feature descriptor learned using both modalities and therefore using them together helps in learning a more discriminative 
feature descriptor.

\section{Conclusions}
In this work, we propose a local feature descriptor for 3D LiDAR scans and a metric for matching the descriptors. We use an architecture based on dense blocks for feature learning and show how our architecture learns more discriminative feature descriptors in comparison to descriptors learned using other common architectures. We report results on matching accuracy and the alignment error. For both cases our learned descriptor outperforms handcrafted descriptors by a significant margin. We also report results for data collected from a different sensor and demonstrate how our descriptor can generalize to different sources of data. We compare the performance of a descriptor learned using a predefined metric and learned with a metric and show that using former is suitable when faster matching time is necessary. Additionally, we also present an ablation study to understand the importance of depth and intensity modalities and show that using them together enables learning of a more discriminative feature descriptor.

\bibliographystyle{plainnat}
\footnotesize
\bibliography{dewan18icra}

\end{document}